# Real-time Prediction for Mechanical Ventilation in COVID-19 Patients using A Multi-task Gaussian Process Multi-objective Self-attention Network

Kai Zhang [a*], Siddharth Karanth [b], Bela Patel [b], Robert Murphy [a], Xiaoqian Jiang [a],

[a] School of Biomedical Informatics, University of Texas Health Science Center at Houston, Houston, TX 77030, USA
[b] Department of Internal Medicine, McGovern Medical School of The University of Texas Health Science Center at Houston, Houston, TX 77030, USA

## Abstract

**Goal:** This paper proposes a robust in-time predictor for in-hospital COVID-19 patient's probability of requiring mechanical ventilation. The data-driven model outputs a highly consistent and robust risk score trajectory for a patient with COVID-19. The patient's risk score at a particular time point indicates the risk of the patient's condition worsening to the point of requiring mechanical ventilation. The model could serve as an early warning monitor system for in-hospital patients. The successful and accurate prediction of such risk scores could help physicians to provide earlier respiratory support for the patient and reduce mortality.

**Methods:** A challenge in the risk prediction of COVID-19 patients lies in the great variability and irregular sampling of patient's vitals and labs observed in the clinical setting. Existing methods have strong limitations in handling time-dependent features' complex dynamics, either oversimplifying temporal data with summary statistics that lose information or over-engineering features that lead to less robust outcomes. We propose a novel in-time risk trajectory predictive model to handle the irregular sampling rate in the data, which follows the dynamics of risk of intubation for individual patients. The model incorporates the Multi-task Gaussian Process (MGP) using observed values to learn the posterior joint multivariant conditional probability and infer the missing values on a unified time grid. The temporal imputed data is fed into a multi-objective self-attention network for the prediction task. A novel positional encoding layer is proposed and added to the network for producing in-time predictions. The positional layer outputs a risk score at each user-defined time point during the entire hospital stay of an inpatient. We frame the prediction task into a multi-objective learning framework, and the risk scores at all time points are optimized altogether, which adds robustness and consistency to the risk score trajectory prediction. The parameters of the MGP are jointly optimized with the downstream multi-objective self-attention network.

**Result:** The model produces in-time robust risk score predictions for each patient – a consistently ascending risk score trend for each patient who would perform mechanical ventilation in the following days and a descending trend for those who would not. On

E-mail addresses: kai.zhang.1@uth.tmc.edu (K. Zhang), Xiaoqian.jiang@uth.tmc.edu (X. Jiang), Siddharth.Karanth@uth.tmc.edu (S. Karanth), Bela.Patel@uth.tmc.edu (B. Patel), Robert.Murphy@uth.tmc.edu (R. Murphy)
* Corresponding author

the other hand, conventional models' risk score trajectory predictions fluctuate and often generate self-conflicting results (e.g., the risk at discharge is higher than the risk 12 hours ago). Our experimental evaluation on a large database with nationwide in-hospital patients with COVID-19 also demonstrates that it improved the state-of-the-art performance in terms of AUC (Area Under the receiver operating characteristic Curve) and AUPRC (Area Under the Precision-Recall Curve) performance metrics, especially at early times after hospital admission. We also evaluate the feature importance of each input feature contributing to the final output.

**Discussion:** The ability to distinguish potentially deteriorating patients from the rest also facilitates reasonable allocation of medical resources, such as ventilators, hospital beds, and human resources. We believe there is a need to develop novel risk prediction models to serve as early warning systems to benefit both in-hospital patients and physicians. Our model makes full use of the in-hospital patient's temporal physiological data, including laboratory tests and vital signs, drug prescriptions, and patient demographic information, etc., to generate robust outcomes.

**Conclusion:** The availability of large and real-time clinical data calls for new methods to make the best use of them for real-time patient risk prediction. It is not ideal for simplifying the data for traditional methods or making unrealistic assumptions that deviate from observation's true dynamics. We demonstrate a pilot effort to harmonize cross-sectional and longitudinal information with multi-objective targets.

## 1. Introduction

The novel coronavirus disease (COVID-19) is caused by the severe acute respiratory syndrome coronavirus 2 (SARS-CoV-2). Today, over 96 million people have been affected, and the virus directly causes more than 2 million deaths worldwide. Among the patients with COVID-19, some people's situations have developed into critical illness and would require respiratory support in their disease course, especially the elders and those patients who have comorbid health conditions [1], [2]. Mechanical ventilation is a crucial medical procedure to help the body maintain healthy oxygen and CO2 level. Therefore, it is usually considered a timely intervention to mitigate patient's condition deterioration. However, after the pandemic outbreak, many countries have experienced a critical situation, that the demand for ventilators and other intensive treatments far outstrips the supply. Clinicians and researchers have developed different workarounds, such as exploring the possibility of sharing ventilators among multiple patients [3], [4]. Therefore, a method for accurate and early recognition of those at-risk patients who will need to perform mechanical ventilation in the future becomes critical for allocating scarce medical resources during the outbreak of a pandemic. Moreover, earlier triage of the patients with a higher risk of being critically ill also enables clinicians to intervene early and apply aggressive treatment to increase the patient's survival rate, which further emphasizes the need for an in-time and accurate prediction model. Unfortunately, the prediction task mentioned above remains an unrealistic task even for experienced clinicians. Many clinical conditions can cause symptoms that may directly or indirectly lead the patients to situations that require a mechanical ventilator. This paper explores the possibility of using data-driven approaches to solve this problem by proposing a neural-network-based prediction model that utilizes the high dimensional physiological temporal data that are generally available for in-hospital patients nowadays. The model could capture the complex interactions among the physiological variables and learn the disease progression pattern to identify patients who need early access to respiratory support and help clinicians perform earlier therapeutic intervention or make resource allocation days ahead of these needs. The ultimate goal is to: 1. Enable physicians to perform close monitoring and timely administration for high-risk patients and reduce the possibility of developing into severe conditions; 2. Help hospitals to allocate scarce critical care resources during the pandemic.

Electronic Health Records (EHRs) are now widely used by physicians and researchers to improve health care quality [5]. The EHR data, especially the longitudinal data, provides an unprecedented opportunity for physicians and scientific researchers to explore and build diagnostic or predictive models, opening the gate of using data-driven approaches to address the healthcare community's inherent challenges. Since the COVID-19 pandemic outbreak, tens of studies on prediction models using EHR data have been proposed while this number is still rapidly growing. For instance, Li et al. proposed a COVID-19 detection neural network (COVNet) by extracting visual features from volumetric chest CT scans to detect COVID-19 [6]. Khan et al. used a deep convolutional neural network model to detect COVID-19 infection from chest X-ray images automatically [7]. Other deep-learning models have also been proposed, which use X-ray images for COVID-19 diagnosis [7]–[16]. Gao et al. proposed a machine-learning-based early warning system to predict the risk of mortality for COVID-19 patients [17]. Babukarthik et al. used genetic deep learning convolutional neural network (GDCNN) to identify and distinguish pneumonia caused by COVID-19 and healthy lungs (normal person) using CXR images [18]. Kapoor et al. proposed a forecasting approach for COVID-19 case prediction using graph neural-networks and mobility data [19]. Douville et al. used Random Forest to predict COVID-19 patient's probability of requiring mechanical ventilation [20]. Alnababteh et al. applied a multiple-regression model on heart rate, SpO2/FIO2 (S/F) ratio, and troponin (TnI) to predict the need for invasive mechanical ventilation [21]. Ferrari et al. used the ensembles of Decision Tree models to predict respiratory failure [22].

Standard parametric and non-parametric models such as logistic regression and tree-based methods perform well on cross-sectional data and are useful for predicting risk score by leveraging the clinical data. Time-to-event analysis such as the semiparametric Cox proportional-hazard model can also be applied to predict case-specific hazards. However, traditional machine learning models view the patient's status as a static variable based on only the most recent observations, thus fall short of taking full advantage of the disease development process. In cases where in-time prediction is favored, these models need to be trained and make predictions at each time point of interest. Considering the patient's condition is time-varying, which depends not only on recent observations but also on the historical observations. Separately training a model at each time point could introduce incoherence into a patient's risk predictions as time passes. The model could achieve good prediction performance on the population level at each time point. However, the risk score predictions at different time points for an individual may vary significantly, and sometimes even conflict over time. The incoherent or highly fluctuating risk scores will bring difficulties when making treatment plans for a patient.

The purpose of this study is to propose a predictive model that could provide timely, consistent, and robust predictions for the probability of performing mechanical ventilation for patients with COVID-19. Overall, we face and address the following challenges in this work. First, typical classification models such as logistic regression, tree-based methods, Cox proportional-hazards model, etc., cannot provide in-time risk predictions despite their high accuracy. Whereas treating each time point separately by training separate models at different time points often produces a good overall performance on the population level, however, on the individual level, the risk predictions could highly fluctuate or even be inconsistent over time. Second, EHR data are often collected in an unscheduled manner such that traditional frameworks usually view most patients as having missing data, which deteriorates the model's performance. Third, the EHR data among different patient encounters often face the problems of asynchronously sampling (lab tests and vital signs are sampled at different timestamps) and irregularly sampling (the time interval between every two contiguous observations are not always consistent), and EHR data often demonstrates a mixture of both patterns which causes the conventional data imputation methods such multiple imputation methods to fail. Fourth, the need for mechanical ventilation (indicating deterioration of the patient's situation) is determined by a combination of many factors whose pattern is too complex for human-beings to capture. A unified scoring system such as the sequential organ failure assessment (SOFA) score is overly simplistic to model the complicated interactions among the observational variables (laboratory, vital signs, demographics, medicines, etc.), and the performance is limited for this task. More advanced models are needed to capture complex intra-series and inter-series relationships among the multivariant longitudinal data. Finally, it is crucial but also difficult to distinguish the patients who would need mechanical ventilation from the others at very early times after admission, since the biochemical indicators only become abnormal after a certain amount of time for most patients, i.e., several hours or even a few days before patient's condition deteriorates and mechanical ventilation is needed.

The proposed model for predicting patient risks addressed these challenges by leveraging the MGP [23] to impute the patient's multivariant longitudinal data with missing values and align the imputed values onto a unified regular time grid. We propose a novel positional encoding layer as the last layer of the neural network, which outputs a risk score at each time point. The risk scores at all time points are optimized jointly by framing the prediction task as a multi-objective learning task. The purpose of the final encoding layer is to ensure the risk score at each time point uses only the patient's observational data beforehand, which brings benefits to the risk score predictions to be more robust and consistent over time without enforcing this behavior in the loss function. We integrate the MGP with a deep neural network to build an end-to-end model that uses a self-attention mechanism to capture complex longitudinal high-dimensional data's latent structure. The deep neural network adopts the attention mechanism which enables the network to 'look back' at the patient's entire observational trajectory and selectively connects two distant observations, thus avoids the time-consuming recursion in typical RNN (Recurrent neural network) models and be more capable of handling long-term dependencies and produce more accurate results. Overall, the contributions of our work are:

1. **Real-time, Robust, Consistent Risk Prediction**. The proposed model predicts the patient's risk in a timely manner. The model produces a risk score upon the patient's admission to the hospital and updates the risk score as more observational data is collected and fed into the model, whereas traditional models often produce high fluctuating or even conflicting risk scores over time. We propose a positional encoding feed-forward layer to frame the problem as a multi-objective learning problem, which in turn benefits the prediction results to be robust and consistent overtime for long sequential data.

2. **Novel Individual-level Evaluation Metrics**. Traditional models optimize the prediction performance at each time separately to achieve overall good performance on the population level but fall short on optimizing individual-level performance. We propose two evaluation metrics to measure the performance of the patient's risk score trend predictions. The model has been shown to produce predictions over time that have both high consistency and robustness.

3. **End-to-end Learning Model.** The MGP and the neural network parameters are trained together rather than separately using backpropagation, which benefits the model's training speed to find the global optimum instead of being trapped into the loss function's local minimum.

Our model is trained on a dataset of patients with COVID-19, and the patient information is derived from the Optum® de-identified COVID-19 Electronic Health Record dataset (2007-2020). We performed screening and select patients whose COVID-19 tests are positive and hospitalized since the pandemic outbreak during the year 2020. We compared our model with traditional prediction models, including logistic regression, tree-based models, and Cox proportional-hazards model, and our model demonstrates the ability to provide in-time, highly consistent, and robust risk score predictions. The model also shows an overall performance improvement when comparing with several neural-network-based models, such as RNN and TCN (Temporal Convolutional Networks) combined with missing data imputing methods.

## 2. Related Works

There has been a large body of works focusing on predicting the need for mechanical ventilation for COVID-19 positive patients since the breakout of the COVID-19 pandemic. For instance, Khandelwal et al. proposed a scoring system named "COVID-19 Score" for predicting the likelihood that patients will require tracheal intubation [24]. Burdick et al. use the XGBoost classifier to fit boosted decision trees on the patient's two-hour data after hospital admission to predict respiratory decompensation in patients with COVID-19 within the next 24 hours [25]. Several risk factors were identified to be associated with intubation and prolonged intubation in hospitalized patients with COVID-19 [26]. They studied time-to-extubating for in-hospitalized patients using multivariable logistic regression analysis. Roca et al. used the ROX index (defined as the ratio of oxygen saturation, measured by pulse oximetry/FiO2) to predict high-flow nasal cannula (HFNC) outcome, i.e., need or not for intubation [27]. An unsupervised symptom time series clustering model was proposed to predict disease severity or the need for dedicated medical support for COVID-19 positive patients [28]. Su et al. used the patient's post-intubation trajectory of SOFA (Sequential Organ Failure Assessment) score to identify and characterize distinct sub-phenotypes of COVID-19 critical illness and classified them into mild, intermediate, and severe groups [29]. Liang et al. used a deep learning-based survival model to predict the risk of patients with COVID-19 developing critical illness based on clinical characteristics at admission using ten selected features by the LASSO algorithm [30]. The variables considered in their study include X-ray abnormalities, age, dyspnea, COPD (chronic obstructive pulmonary disease), number of comorbidities, cancer history, neutrophil/lymphocytes ratio, lactate dehydrogenase, direct bilirubin, and creatine kinase.

These models successfully achieve the prediction goal but with certain limitations. The majority of the models cannot provide consistent in-time risk predictions over time due to the lacking of a Spatio-temporal attention mechanism by optimizing the model's performance at each time point individually. Some models use temporal features; however, these features are overly simplistic and do not take full advantage of all available patient information. For example, the SOFA score is defined only on six most important features: PaO2, FiO2, platelets, bilirubin, mean arterial pressure, creatinine, and the ROX score is defined on only three factors: PaO2, FiO2, respiratory rate. The feature-selection based (such as LASSO) models may induce information loss, especially when the covariant are correlated, which leads to lower accuracy results. Moreover, the patient's previous observations' time plays an important role in future disease development. Many of the above temporal models fail to capture the complex patterns in the multivariant temporal trajectories and long dependencies.

The challenge we face when leveraging longitudinal EHR data for prediction tasks is that the data is often collected irregularly -- lab tests are rarely collected on a fixed routine, and vital sign observations are missing due to the patient leaving, etc., which cause patient records as unstructured. The natural heterogeneous inpatient encounters in the dataset often render many traditional learning frameworks viewing the data as has missing values. Multiple Imputation (MI) [28] is a widely used approach to impute missing data in longitudinal studies. Multiple imputation techniques can be classified into two major categories based on the assumptions of the underlying distributions that the data follows. The fully conditioned specification approach imputes missing data based on an estimated univariant posterior distribution of the random variable given all other variables and iteratively do this for all other random variables with missing values [31]; the jointly modeling approach is based on the assumption that the random variables are jointly Gaussian [32]. The above multiple imputation methods cannot be directly used, and a workaround is adopted by usually treating measurements of one variable at different times as separate variables [33]. Recently, researchers have been working on tailoring the above methods to longitudinal data, such as the generalized linear mixed-effects model (GLMM)-based methods [31], [33]–[37], which demonstrate to provide more accuracy compared to standard MI approaches.

The multiple imputation method falls short on handling the irregularity in longitudinal data. To this end, Baytas et al. proposed a novel LSTM architecture (T-LSTM) to handle irregularities in time sequences for patient subtyping [38]. They used ICD-9 codes as a one-hot representation to feed into the LSTM model for prediction tasks by adding the time intervals between consecutive hospital visits as another dimension of the model input. Tan et al. provided an end-to-end dual-attention time-aware gated recurrent unit (DATA-GRU) for predicting patient mortality risk [39]. Bahadori et al. proposed a multi-resolution ensemble (MRE) model, which is an ensemble method for predictions by applying different coarsening operators on time-sequence data [40]. A deep learning model based on GRU, named GRU-D, was developed by [41] to capture the long-term temporal dependencies by assimilating the masking and time interval patterns into the deep learning architecture. Bokde et al. proposed a method called "imputePSF" that facilitates the sequential data's repeating characteristics as a statistical basis for the imputation task [42]. However, they only focus on univariate time series data. Recently, Choi et al. provided a training approach named RDIS (Random Drop Imputation with Self-Training) [43]. The idea is to introduce extra missing values by applying a random drop on the already incomplete data, and the neural network is trained to predict the pseudo-missing values correctly. There is a substantial amount of deep learning-based imputation methods besides the above, and a summary of such works is given by [44].

This study uses the Multi-task Gaussian Process to model multivariant time-series data, which has been shown to have significant success for temporal missing data imputation on similar tasks. Dürichen et al. used multi-task Gaussian processes [23] to perform multivariate physiological time-series analysis [45]. Futoma et al. combined the multi-task Gaussian process and the RNN network as a joint model for imputation-prediction task, which showed performance improvement on predicting Sepsis compared to various previous pipelines [46]. Futoma et al. proposed a multi-output Gaussian process deep recurrent Q-network [47]. Moor et al used Gaussian process and temporal convolutional networks (TCN) and dynamic time warping (DTW) for early recognition of Sepsis [48]. Innovated from the above works, our model leverages the Multi-task

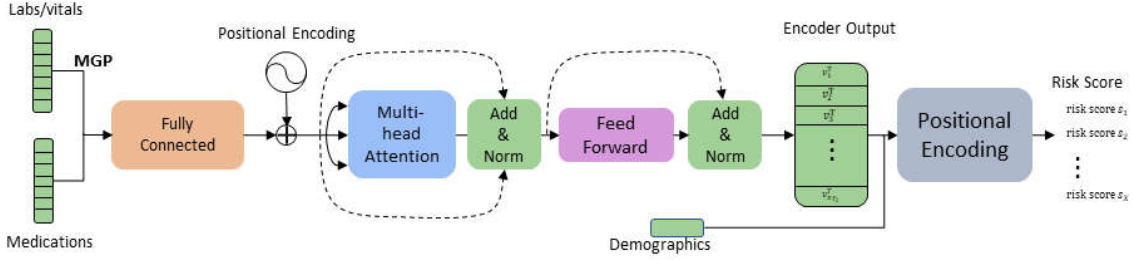

Figure 1. MGP-MS Model Overview. The model combines a Multi-task Gaussian Process module with the multi-objective attention mechanism for trajectory prediction.

Gaussian Process combined with an improved self-attention neural network to impute the missing values in the multivariant longitudinal EHR data.

## 3. Methods

The purpose of this study is to provide a predictive model that overcomes the weaknesses of existing models by incorporating patient's most available data after hospital admission, including physiological data, demographical information, and drug prescriptions, to produce more timely, consistent, and robust predictions for each individual.

In this section, we describe the multi-task Gaussian process multi-objective self-attention (MGP-MS) pipeline. Section 3.1 provides a background of the multi-task Gaussian process used as a missing data imputation mechanism in our task. Section 3.2 introduces the multi-objective predictive model using a multi-objective self-attention neural network. Section 3.3 explains the data filtering process, and section 3.4 introduces the feature importance evaluation.

### 3.1. Multi-task Gaussian Process

We use the Multi-task Gaussian Process to impute the missing values in multivariate time series data, specifically, the missing values of lab tests and vital signs, where each lab or vital can be viewed as one task. The time-lapse between consecutive observations of lab test and vital sign of hospitalized COVID-19 patients often have irregular property. For each patient $i$, let $T_i$ be the number of observational times during patient $i$'s hospital stay. We denote the observed lab tests and vital signs as $(\boldsymbol{t}_i, \boldsymbol{Y}_i)$, where the $\boldsymbol{t}_i = [t_{i1}, \ldots t_{iT_i}]$ is a vector of observational time points (not necessarily even-spaced) and $\boldsymbol{Y}_i = [\boldsymbol{y}_{i1}, \ldots, \boldsymbol{y}_{iT_i}], \boldsymbol{y}_{ij} \in \mathbb{R}^D$ is a vector of observed values at those time points. Function $f_{id}(t)$ is used to represent the latent function of time $t$ for patient $i$'s $d$-th feature. In our task, we assume the prior distribution of the Gaussian process has zero mean value, and we have

$$\langle f_{ij}(t), f_{ij'}(t') \rangle = K_{jj'}^f k^t(t, t'), \tag{1}$$

$$y_{id}(t) \sim \mathcal{N}(f_{id}(t), \sigma_d^2), \tag{2}$$

where $y_{id}(t)$ is the observed value of variable $d$ at time point $t$ of the patient $i$. Similarly, if we denote $\boldsymbol{y}_i \triangleq vec(\boldsymbol{Y}_i) = [y_{i,11}, \ldots, y_{i,T_i 1}, y_{i,12}, \ldots, y_{i,T_i 2}, \ldots, y_{i,1D}, \ldots, y_{i,T_i D}]$, then $\boldsymbol{y}_i$ follows the distribution

$$\boldsymbol{y}_i \sim \mathcal{N}(\boldsymbol{0}, K^D \otimes K^{T_i} + E \otimes I), \tag{3}$$

where $K^D$ is a $D \times D$ matrix representing the inter-task similarities, and $K^{T_i}$ is a kernel matrix representing the similarities among the observational times. In this work, we use the squared exponential kernel function $k_{\text{SE}}(t, t') = \exp(-|t - t'|^2/(2l^2))$, where $l$ is the length scale of the process. The diagonal matrix $E$ denotes the noise variances, $E = \text{diag}(\sigma_d^2), d = 1, \ldots D$.

The purpose of MGP is to use all the observations $\boldsymbol{y}_i$ available to us to impute the values on a regularly spaced time grid $\boldsymbol{x}_i = \{x_{i,1}, x_{i,2}, \ldots, x_{i,X_i}\}$, where $|\boldsymbol{x}_i| = X_i$. We denote the imputed values on the grid as a vector of $\boldsymbol{z}_i$, and the goal of MGP is to estimate the posterior distribution $P(\boldsymbol{z}_i | \boldsymbol{y}_i, \boldsymbol{t}_i, \boldsymbol{x}; \boldsymbol{\theta})$, where $\boldsymbol{\theta}$ is the parameter(s) of the multi-task Gaussian process. The $\boldsymbol{z}_i$ values have the following distribution

$$\boldsymbol{z}_i \sim \mathcal{N}(\boldsymbol{\mu}(\boldsymbol{z}_i), \boldsymbol{\Sigma}(\boldsymbol{z}_i); \boldsymbol{\theta}), \tag{4}$$

$$\boldsymbol{\mu}(\boldsymbol{z}_i) = (K^D \otimes K^{XT_i}) \Sigma_i^{-1} \boldsymbol{y}_i, \tag{5}$$

$$\boldsymbol{\Sigma}(\boldsymbol{z}_i) = (K^D \otimes K^X) - (K^D \otimes K^{XT_i}) \Sigma_i^{-1} (K^D \otimes K^{T_i X}), \tag{6}$$

where $K^{XT_i}$ is the kernel matrix among all observational time points and the regularly spaced time grids. $K^X$ is the kernel matrix denoting the similarity among regularly spaced times $\boldsymbol{x}$. The $\otimes$ denotes the Kronecker product. The parameter $\boldsymbol{\theta}$ denotes all parameters of the Gaussian process to be learned, i.e., $\boldsymbol{\theta} = \{K^M, E, l\}$. The multi-task Gaussian process outputs a $X_i \times D$ matrix $\boldsymbol{z}_i$, which includes the observational values of the $D$ lab tests and vital signs on the time grid $\boldsymbol{x}$ having a regular format to input to the following attention-based neural network.

### 3.2. Multi-objective Optimization

The task of predicting the risk of performing mechanical ventilation is realized by a multi-objective self-attention network. Specifically, we leverage only the transformer-encoder network [49], which takes the imputed data $\boldsymbol{z}_i$ and outputs a risk score $s_{i,j}$ at each time point $x_{i,j}, j = 1, \ldots, X_i$. The post-imputation labs and vital data $\boldsymbol{z}_i$ will be concatenated with the medicine prescriptions data $\boldsymbol{m}_i$ before feeding into the neural network. Let $M$ be the total number of medicines we are interested in, then $\boldsymbol{m}_i$ is a matrix of size $X_i \times M$ and in our dataset is a one-hot representation of patient $i$'s medical prescriptions at each time point $x_i$. The data $(\boldsymbol{z}_i, \boldsymbol{m}_i)$ is the final input to be fed into the multi-objective self-attention network. First, each row of the matrix $(\boldsymbol{z}_i, \boldsymbol{m}_i)$, a vector of length $(D + M)$, is encoded to another space (length-$E$ vector) using an embedding layer. The encoder network consists of a self-attention module and a feed-forward module and also incorporates a positional encoding mechanism in the front-end. Several encoders are stacked together for complicated tasks to ensure performance, and the first encoder incorporates the positional information and the embeddings as input. For the self-attention module and a feed-forward module, we refer the readers to the landmark study [49], and here will omit the details. We denote the output of the final layer as $V_i$, which is an $X_i \times E$ matrix.

A block-wise upper-triangular positional encoding layer is designed to ensure that a score at a time point is predicted using only the information before that time. The final encoder's output $V_i$ will be flattened row-wisely to a vector of length $X_i E$ and concatenated with the patient's baseline covariance information $w_i$ before multiplying with the positional encoding layer. The final risk score output will be a length $X_i$ vector of real values, representing the score of risk at each time,

$$(v_{i1}^T, \ldots, v_{iX_i}^T, w_i^T) \bullet \begin{pmatrix} B_1 & B_1 & \cdots & B_1 \\ 0 & B_2 & \cdots & B_2 \\ 0 & 0 & \cdots & B_3 \\ \vdots & \vdots & \ddots & \vdots \\ 0 & 0 & \cdots & B_{X_i} \\ P & P & \cdots & P \end{pmatrix}$$
$$= (s_{i1}, \ldots, s_{iX_i}) \tag{7}$$

where $w_i^T$ is a vector of dimension $F$ representing the demographics data of patient $i$, each column vector $B_j, j = 1, \ldots, X_i$ is of dimension $E$, and $P$ is a column vector of dimension $F$. The proposed model outputs a risk score $s_{ij}$ at each time of $x_{i,j}$ during the patient $i$'s hospital stay length.

The MGP together with the multi-objective self-attention neural network can be viewed as an implicit function $h(t_i, y_i, x, m_i, w_i; \theta, \omega)$, where $\theta, \omega$ are the MGP parameters and weights in the neural network, respectively. The loss function of the proposed model is defined as

$$\theta^*, \omega^* = \operatorname{argmin}_{\theta, \omega} \sum_{i=1}^{N} \sum_{j=1}^{X_i} \mathrm{E}_{z_i \sim \mathcal{N}(\mu(z_i), \Sigma(z_i); \theta)} l(s_{ij}, o_{ij}), \tag{8}$$

where $N$ is the total number of patients in the training dataset and $l$ is the loss function where we use the cross-entropy function. The true label $o_i$ of a patient $i$ is replicated $X_i$ times to form the multi-objective optimization problem, i.e. $o_i = o_{ij}, \forall j$.

The model's prediction's consistency and robustness are ensured by the last block-wise upper-triangular positional encoding layer. By letting the model 'foresee' the outcome at each previous time point ahead of the event (performing mechanical ventilation) occurs, the model is enforced to predict the correct outcome at all time points rather than being correct only at the end of the observational trajectory. Introducing the multi-objective mechanism brings consistency and robustness to the prediction results such that the high-risk and low-risk patients have distinct (ascending or descending) risk score pathways.

### 3.3. Data Truncation

We perform data truncation to select only the valid patient's data to feed into our model. Specifically, a left truncation is performed to exclude the patients who performed mechanical ventilation by the time the patient was admitted to the hospital. It is critical to be able to identify the high-risk patients in a fairly short amount of time after admissions. Therefore, we set the study period to be, for example, 24 hours. We also exclude all patients from our dataset who stayed in the hospital for less than 24 hours and those patients who performed mechanical ventilation in less than 24 hours. We will use the first 24-hour observational data after admission to learn a temporal model and use it to predict the 24-hour risk score trajectory for an unseen patient.

### 3.4. Feature Importance Estimation

The feature importance can be measured by the amount of performance decrease of a well-trained model by manipulating the feature of interest while keeping the other features unmodified. Recently, experiments from [50] show that permutation-based feature importance methods could result in unreliable results, especially when the covariant are correlated. In this work, we adopted an alternative suggested in their work by evaluating a newly trained model's accuracy drop after removing a specific feature.

## 4. Experiments

### 4.1. Dataset

We trained and evaluated our model on a cohort of 9,532 hospitalized patients whose COVID-19 test is positive. We set the study period to be the first three days (72 hours) after hospital admission. That is, we would like to train a model to accurately predict a patient's risk score at each time point during the first 72 hours after admission. After an initial data cleansing to remove the features that have too few observations, we selected 16 lab tests and 9 vital signs where each feature should be measured by >50% of all patients at least twice. The data completeness of the selected 25 features is shown in Figure 2. The medicine prescription data is also formulated as sequential data, and we group different medicine names into 21 medicine categories. Finally, the patient's demographic data, including race, ethnicity (encoded as one-hot representation), age, gender (25 variables in total), are concatenated with the attention model's result and fed into the final feed-forward layer. The lab tests, vital signs, and medicine prescription's timestamps are measured in seconds, and we use the 4-hour average value, i.e., take average results in every consecutive 4 hours since admission. After preprocessing, among the 9,532 patients, 1,485 (15.58%) patients have performed mechanical ventilation after the 3 days since admission.

### 4.2. Experiment Setup

We split our dataset into a training set (80%) and a testing set (20%). We set the period of study to be the first 72 hours after admission using a 4-hour window average. Thus the patient $i$'s regularly spaced time grid is $x_i = \{x_{i,j}\}, j = 1, \ldots, 17$ where each $x_{i,j}$ denotes the $j$-th 4-hour window after hospital admission.

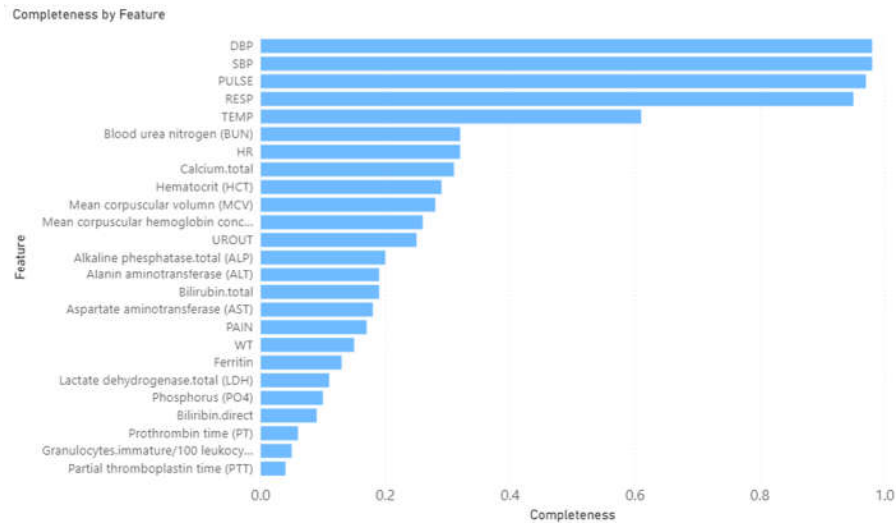

Figure 2. Data Completeness of Lab Tests and Vitals. 100%: feature's data is fully complete.

We compared our model with conventional classification models and survival models, including logistic regression, tree-based methods, and the Cox proportional hazards model. Since the traditional models do not produce in-time predictions, to ensure a fair comparison, we train the above machine learning models on patient data at the following timestamps: upon admission, $4^{th}$ hour, $8^{th}$ hour, $12^{th}$ hour, ..., and the $72^{nd}$ hour after admission. Each model will predict a patient's risk score at the above time points, therefore a trajectory. A smooth and consistent risk score pathway prediction is deemed to be a robust prediction instead of one that highly fluctuates and produces inconsistent predictions. We evaluate and compare the robustness of the proposed model with traditional methods.

Several baseline models were proposed for prediction tasks using irregularly sampled data, including the standard LSTM [51] and GRU [52] neural network, the time-aware LSTM network (T-LSTM) [38], the GRU-D [41], the Gaussian process temporal convolutional networks (GP-TCN) [48], the Interpolation-prediction networks (IPN) [53] and the multi-objective Gaussian process RNN model (MGP-RNN) [46]. This section compares them with the proposed model in terms of the AUC and AUPRC performance metrics.

Our model's hyperparameter is set to have an embedding size of 512; the feed-forward network dimension is 2048; we use 6 encoder layers and 8 attention heads in the multi-objective attention module. The L2 regularization and a dropout rate of 0.3 are added to avoid model overfitting. We use the Adam optimizer [54], and a learning rate scheduler is adapted to adjust the learning rate (starting from 0.03) based on the number of epochs, i.e., decaying the learning rate by a multiplicative factor of 0.95 after each epoch. The model is trained for 100 epochs with a batch size of 50 patients. During the joint-training of the Multi-task Gaussian Process and the neural network, the MGP generates 50 Monte Carlo Samples (pseudo-patients) for each original patient as neural network's input. We implemented our pipeline in PyTorch, and the model is trained on Nvidia Tesla V100 Tensor Core.

### 4.3. Individual-level and Population-level Evaluation Metrics

Based on the goal of the prediction task, we evaluate the model's performance on both the individual and population levels.

**Individual-level:** The model should be able to provide consistent and robust risk score prediction over time for each individual. We propose two performance evaluation metrics to measure the *consistency*, *robustness* of the predicted risk score trajectory.

**Population-level:** The model should be able to distinguish the patients on the whole population who would perform mechanical ventilation from those who would not perform it. We use AUC and AUPRC to evaluate the binary-classification task's overall accuracy on the entire population at each time point.

A patient's risk score trajectory should demonstrate an overall upward tendency for those patients who would perform mechanical ventilation and a downward tendency for those who would not. Therefore, we fit the risk score trend using a linear function and use the function slope to measure consistency. To be precise, we define the *consistency* to be the absolute value of the slope, since the descending trend has a negative slope. Hence, large consistency means a more obvious development tendency (positive or negative) and clearer trend.

The risk score trend should also be robust and not frequently fluctuating significantly over time. Therefore, we first measure the sum squared error between the linear fitting function and the real risk score, then define *robustness* to be (1-error)/(1+error) to constrain the robustness in the range of [0, 1]. Hence, large robustness means less fluctuation of the risk score trend.

We fit the risk score trend of a patient $i$, $\boldsymbol{s}_i = (s_{i1}, \ldots, s_{iX_i})$ at time points $\boldsymbol{x}_i = \{x_{i,1}, x_{i,2}, \ldots, x_{i,X_i}\}$ using linear regression. Suppose the fitted function is denoted as $h(\boldsymbol{x}_i)$, an individual $i$'s risk score trend consistency and robustness are defined as

$$\text{consistency}_i \triangleq |\text{slope}(h(\boldsymbol{x}_i))|, \tag{9}$$

$$\text{robustness}_i \triangleq \frac{1 - \frac{1}{X_i}\sum_{j=1}^{X_i}(h(x_{i,j}) - s_{ij})^2}{1 + \frac{1}{X_i}\sum_{j=1}^{X_i}(h(x_{i,j}) - s_{ij})^2}, \tag{10}$$

respectively.

### 4.4. Results

We compare our model with standard classification and regression models, including Logistic Regression, Gradient Boosted Tree Model (using XGBoost), and Cox Proportional-Hazard Model. Since these models cannot output consistent in-time predictions, each model is trained using the instant physiological data collected during a 4-hour window and makes risk score predictions every 4 hours.

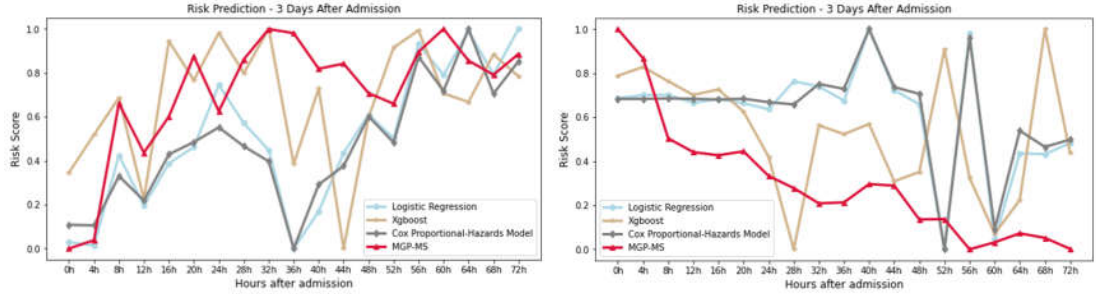

Figure 3 Risk Score Trajectory Prediction Using Different Models. The four colored lines correspond to Logistic Regression, XGBoost, Cox Proportional-Hazard Model, and the proposed MGP-MS model. Left figure: the risk score pathway of a patient with COVID-19 who would perform mechanical ventilation after the 3 days since admission, where each line represents a model's result on this patient. Right figure: the risk score pathway of a patient with COVID-19 who would not perform mechanical ventilation after the 3 days since admission. The proposed method produces consistent and clear ascending and descending risk score progression pathways for the two patients.

The risk score is defined in different ways in a different model. For Logistic regression, XGBoost, and the proposed MGP-MS model, we frame the prediction task as a classification problem, and the final goal is to accurately distinguish the patients who will perform mechanical ventilation from those who would not. Accordingly, the risk score is defined to be the log probability of the model output. For Cox Proportional Hazard Model, we set the task as a regression problem and regress the predictor variables to the survival time (time between admission and mechanical ventilation), and the risk score is defined to be the partial hazard of the individual.

**Individual-level:**

For a particular patient, the risk score prediction among different models is not the same, even at the same time point. Therefore, these risk values produced by different models are not comparable, and here we focus on the tendency of the risk score trajectory rather than the particular values themselves. To this end, we normalize all models' predictions to the range 0.0-1.0. In Figure 3, the left figure shows the risk score prediction of a randomly selected patient in the dataset who would perform mechanical ventilation after 3 days, and the right figure shows that of a patient who would not after 3 days. The four risk score trajectories correspond to the prediction using four different models. The proposed MGP-MS model produces an obvious ascending risk score trend (red-triangle) for the patient who would perform mechanical ventilation and a descending trend for the other one. On the other hand, the other three compared models cannot make consistent predictions over time and the risk score pathway is relatively unstable.

We evaluated each patient's 3-day risk score trend consistency (in Figure 4) and robustness (in Figure 5) predicted by the four models. It can be seen our proposed model's consistency improves on the other models by around 225.00%, 160.00%, and 313.00%, respectively. The proposed model's robustness improves on the other models by around 5.56%, 8.05%, and 6.74%, respectively.

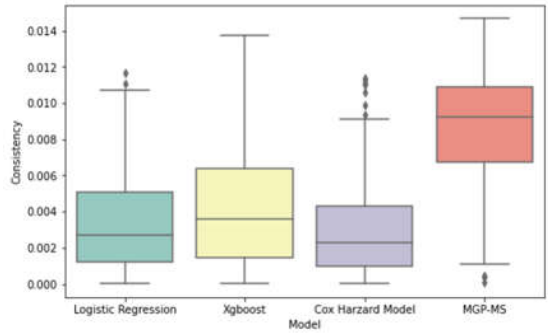

Figure 4 Consistency Evaluation of Different Models.

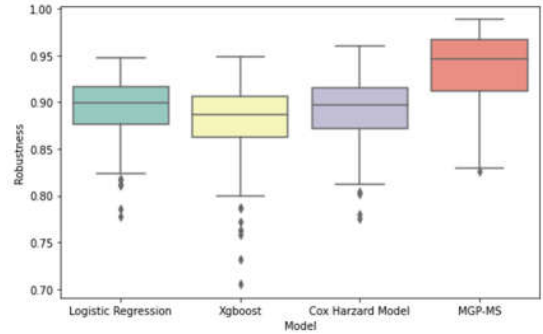

Figure 5 Robustness Evaluation of Different Models.

In Figure 6, each patient's 3-day risk score trajectory's robustness and consistent are shown as a scatter plot where the two evaluation metrics are formed as perpendicular axes. The horizontal axis is the slope of the fitted linear function, and the vertical axis is the robustness. For the two classes of patients, the figure shows the proposed model has more separable slope values for the two categories, where the slope is positive (an increasing risk score trend) for each patient who would need mechanical ventilation and negative (a decreasing risk score trend) for those who would not need it.

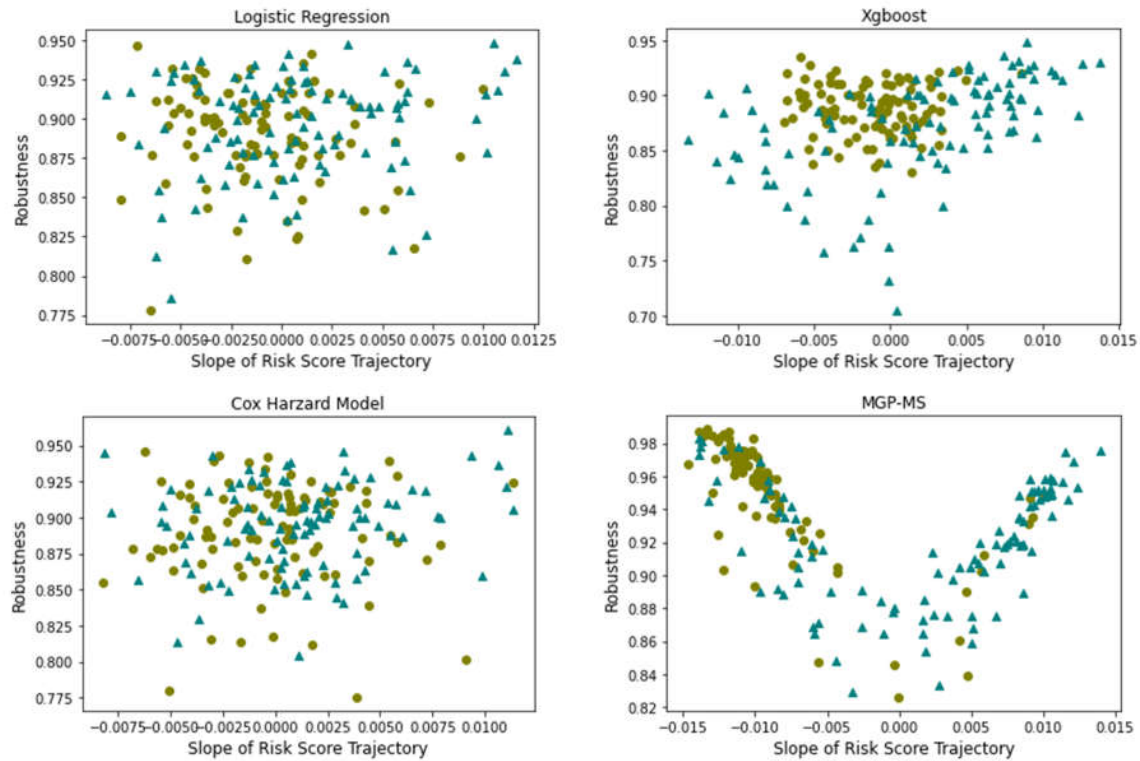

Figure 6 Robustness and Consistency Evaluation of the Models. The four figures represent the consistency - robustness plot of the Risk Score Prediction using Logistic Regression, XGBoost, Cox Proportional-Hazard Model, and the proposed MGP-MS model, respectively. The figure demonstrates the model's consistency (horizontal axis) and robustness (vertical axis). Each orange-triangle or blue-circle represents a COVID-19 patient who would need mechanical ventilation or who would not need it. The x-axis measures the model's consistency, and the y-axis measures the robustness, i.e., the fluctuations of risk score trend. The proposed model shows the proposed model produces high consistency (positive slope trend for patients who would need mechanical ventilation and negative slope trend for those who would not) and relatively low fluctuations (i.e., high robustness) compared to the other three models.

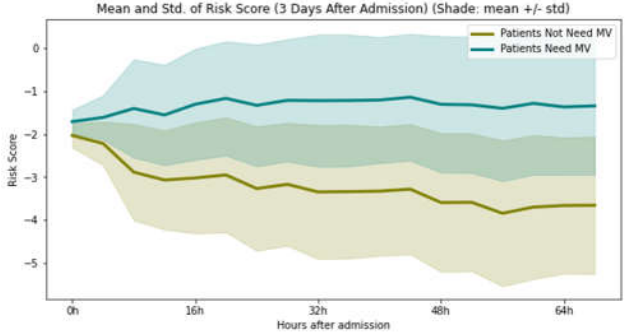

Figure 7. The Average Risk Score Trajectory of the Two Classes of Patients. The shaded area denotes the +/- standard deviation. Light yellow: the average of all patients who will perform mechanical ventilation; Light blue: the average of all patients who will not perform mechanical ventilation.

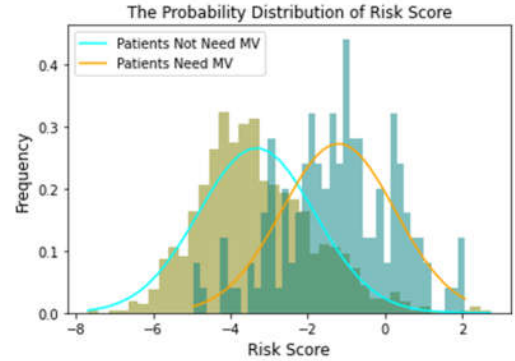

Figure 8. The Risk Score Distribution of The Two Classes of Patients at a particular timestamp. Light blue represents the risk score distribution of all patients who need mechanical ventilation (MV), and light yellow represents those who would not need it. The figure uses the risk score predictions of all patients at the 64th hour after hospital admission.

Figure 7 demonstrates the mean 3-day risk score pathway of patients that would perform mechanical ventilation after 3 days (light blue) and 20 patients that would not (light yellow), and the shaded area is the +/- one standard deviation. The model successfully distinguishes the two classes of patients, i.e. similar pathways within a class and different across the two classes. The patients with a risk score trajectory that resides inside the overlapping area denote patients whose risk trajectory does not have a clear upward or downward trend. This near-flat trajectory denotes the patients whose future situation is uncertain to our model. In reality, physicians can apply different therapeutic interventions for the three distinct pathway patterns. For a patient with uncertainty, we can rely on the human decision or waiting for a longer observation time in the allowance of the hospital resource for the model to collect more evidence to decide.

In Figure 8, we plot the risk score distribution at a specific time point for all patients. The model successfully identifies statistically distinctive risk score distributions for the two classes.

**Population-level:**

On the population level, we summarize the performance of our proposed model and several other models for the mechanical ventilation prediction task on the same dataset in Table 1. The GRU-ffill stands for GRU network using forward-filling when dealing with missing data. The GRU-D, MGP-TCN, IPN, T-LSTM, and MGP-GRU are recently-proposed deep-neural-network-based models that have been shown success for similar tasks. To ensure a fair comparison, we tuned the default hyper-parameters the authors have provided to ensure the best performance, and the predictions at any time point using these models are based on all previous observations since hospital admission. Our model outperforms most of the previous models and gains the largest improvement at the early times after admission.

Table 1. AUC and AUPRC Performance Comparison

| Model | Admission | | 0.5 Day | | 1 Day | | 2 Days | | 3 Days | |
|---|---|---|---|---|---|---|---|---|---|---|
| | AUC | AUPR | AUC | AUPR | AUC | AUPR | AUC | AUPR | AUC | AUPR |
| GRU-ffill | 0.7378±0.0487 | 0.3193±0.0562 | 0.7404±0.0968 | 0.3348±0.0626 | 0.7901±0.0333 | 0.3756±0.0388 | 0.7754±0.0760 | 0.4048±0.0688 | 0.8061±0.0507 | 0.4447±0.0176 |
| GRU-D | 0.6080±0.0064 | 0.2420±0.0094 | 0.6062±0.0077 | 0.2442±0.0096 | 0.6747±0.0093 | 0.3031±0.0072 | 0.7517±0.0107 | 0.4109±0.0292 | 0.8099±0.0055 | **0.4814±0.0182** |
| MGP-TCN | 0.5972±0.0152 | 0.2220±0.0064 | 0.5862±0.0127 | 0.2135±0.0084 | 0.6394±0.0135 | 0.3131±0.0132 | 0.7602±0.0037 | 0.3909±0.0053 | 0.7732±0.0153 | 0.4632±0.0125 |
| IPN | 0.7034±0.0136 | 0.3473±0.0056 | 0.7286±0.084 | 0.3687±0.0043 | 0.7605±0.0094 | 0.3904±0.0148 | 0.7653±0.0158 | 0.4116±0.0053 | 0.7770±0.0198 | 0.4014±0.0098 |
| T-LSTM | 0.5051±0.0020 | 0.1836±0.0098 | 0.5132±0.0066 | 0.2020±0.0293 | 0.5540±0.0046 | 0.2642±0.0099 | 0.6140±0.0156 | 0.3525±0.0054 | 0.6910±0.0045 | 0.3956±0.0124 |
| MGP-GRU | 0.7048±0.0036 | 0.2912±0.0021 | 0.7329±0.0078 | 0.3176±0.0100 | 0.7631±0.0101 | 0.3555±0.0069 | 0.7859±0.0023 | 0.4398±0.0145 | 0.7912±0.0089 | 0.4712±0.0120 |
| MGP-MS | **0.7920±0.0063** | **0.3992±0.0112** | **0.8221±0.0030** | **0.4507±0.0043** | **0.8292±0.0046** | **0.4587±0.0066** | **0.8420±0.0052** | **0.4678±0.0062** | **0.8421±0.0056** | 0.4813±0.0096 |

We evaluated each lab test's feature importance and vital signs, and we show the 15 most important features out of all and illustrated them in Figure 9. The evaluation technique is based on drop-variable-importance proposed in [50], where the increase of prediction error is evaluated by removing each interesting variable and retraining a new model. The vertical axis denotes the increase of the loss after deleting a feature and retrain the model.

## 5. Conclusions

We presented a novel data-driven early warning model to predict the COVID-19 patient's risk score and distinguish the patients that will need mechanical ventilation or not soon after admission. The model provides accurate, robust, and in-time risk score predictions at each time point since admission. We evaluated our model on a cohort of around 10,000 COVID-19 patients and demonstrated high accuracy. We also compared our model performance with several models that have been proved successful for similar tasks and demonstrated a clear performance improvement at each time point. Specifically, the model shows a significant performance improvement at earlier times after admission compared to other models.

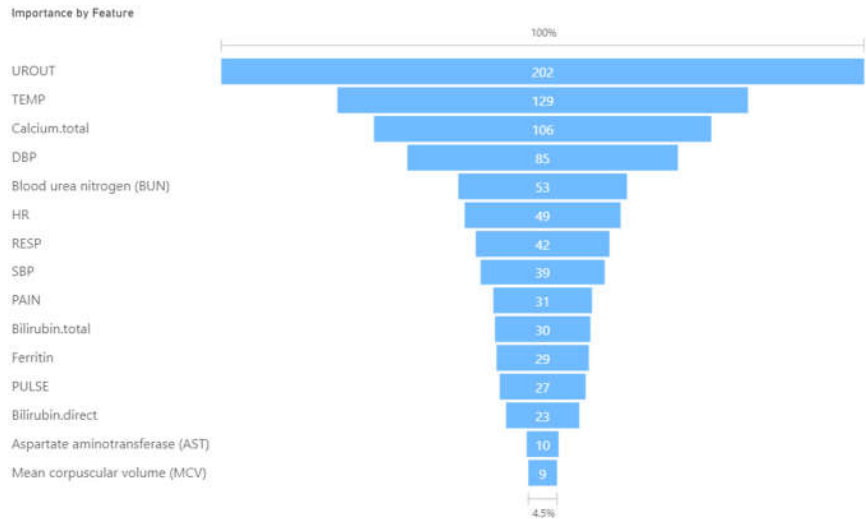

Figure 9. Feature Importance Evaluation. The feature importance evaluated for lab tests and vital signs, and the figure shows the top 15 most important features out of all.

The model has clinical significance during the circumstances of the ongoing COVID-19 pandemic, especially for the hospitals that are experiencing a shortage of ventilators due to the increasing number of in-hospital patients. The allocation of ventilators to the patients of higher risk would significantly increase the overall survival rate.

## Acknowledgments

XJ is CPRIT Scholar in Cancer Research (RR180012), and he was supported in part by Christopher Sarofim Family Professorship, UT Stars award, UTHealth startup, the National Institute of Health (NIH) under award number R01AG066749, R01GM114612 and U01TR002062, and the National Science Foundation (NSF) RAPID #2027790. KZ is supported in part by CPRIT RR180012. KZ is supported by Cancer Research (RR180012).